\documentclass[conference]{IEEEtran}
\IEEEoverridecommandlockouts
\usepackage{cite}
\usepackage{amsmath,amssymb,amsfonts}
\usepackage{algorithmic}
\usepackage{graphicx}
\usepackage{textcomp}
\usepackage{xcolor}
\usepackage{subcaption}
\def\BibTeX{{\rm B\kern-.05em{\sc i\kern-.025em b}\kern-.08em
    T\kern-.1667em\lower.7ex\hbox{E}\kern-.125emX}}
\begin{document}

\title{Training Experimentally Robust and Interpretable Binarized Regression 
Models Using Mixed-Integer Programming
\thanks{Work supported by the Faculty of Information Technology at Monash University 
through the 2021 Early Career Researcher Seed Grant.}
}


\author{\IEEEauthorblockN{Sanjana Tule}
\IEEEauthorblockA{\textit{Monash University}\\
Melbourne, Australia \\
stul0004@student.monash.edu}
\and
\IEEEauthorblockN{Nhi Ha Lan Le}
\IEEEauthorblockA{\textit{Monash University}\\
Melbourne, Australia \\
hlee0047@student.monash.edu}
\and
\IEEEauthorblockN{Buser Say}
\IEEEauthorblockA{\textit{Monash University}\\
Melbourne, Australia \\
buser.say@monash.edu}
}

\maketitle

\begin{abstract}
In this paper, we explore model-based approach to training robust and interpretable binarized 
regression models for multiclass classification tasks using Mixed-Integer Programming (MIP). Our 
MIP model balances the optimization of prediction margin and model size by using a weighted objective 
that: minimizes the total margin of incorrectly classified training instances, maximizes the total 
margin of correctly classified training instances, and maximizes the overall model regularization. We 
conduct two sets of experiments to test the classification accuracy of our MIP model over standard 
and corrupted versions of multiple classification datasets, respectively. In the first set of 
experiments, we show that our MIP model outperforms an equivalent Pseudo-Boolean Optimization 
(PBO) model and achieves competitive results to Logistic Regression (LR) and Gradient Descent (GD) 
in terms of classification accuracy over the standard datasets. In the second set of experiments, 
we show that our MIP model outperforms the other models (i.e., GD and LR) in terms of 
classification accuracy over majority of the corrupted datasets. Finally, we visually demonstrate 
the interpretability of our MIP model in terms of its learned parameters over the MNIST dataset. 
Overall, we show the effectiveness of training robust and interpretable binarized regression models 
using MIP.
\end{abstract}


\section{Introduction}
\label{sec:intro}

Machine learning models have significantly improved the 
ability of autonomous systems to solve challenging tasks, 
such as image recognition~\cite{Krizhevsky2012}, speech 
recognition~\cite{Deng2013} and natural language 
processing~\cite{Collobert2011}. The rapid deployment 
of such models in safety critical systems resulted in 
an increased interest in the development of machine 
learning models that are \textit{robust} and 
\textit{interpretable}~\cite{Rudin2019}. 

In the context of machine learning, robustness refers 
to the performance stability of the model in the presence 
of natural and/or adversarial changes~\cite{Rosenfeld2020}. 
For example, random noise in the training dataset, or random 
changes in the environment can both significantly degrade the 
testing accuracy of a machine learning model when they are not 
accounted for~\cite{Natarajan2013}. 
Orthogonal to robustness, interpretability is concerned with the 
insights that the learned model can provide to its users about the 
relationships that are present in the dataset or the 
model~\cite{Murdoch2019}. Interpretable models are desirable as 
they often drive actions and further discoveries. Interpretable 
machine learning aims to construct models that are low in complexity 
(i.e., typically measured by the size of the model), highly simulatable 
(i.e., how easy for the user to simulate the prediction of the model), 
and highly modular (i.e., can each subcomponent of the model be 
evaluated separately). Interpretable machine learning can be thought 
of as a set of criteria that should be considered prior to the 
model selection.

In this paper, we train interpretable machine learning models 
that are experimentally robust to random corruptions in the dataset. 
Specifically, we focus on performing the supervised multiclass 
classification task, and limit our scope to random corruption of 
labels. Based on the interpretability criteria (i.e., complexity, 
simulatability and modularity), we select binarized Linear 
Regression as the appropriate machine learning model, and optimally 
learn its parameters by solving an equivalent Mixed-Integer 
Programming (MIP) model. We experimentally demonstrate the 
performance benefit of training a binarized Linear Regression model 
using MIP over both standard and corrupted datasets, and visually 
validate the interpretability of the learned model using the MNIST 
dataset. Overall, we contribute to the interpretable machine learning
literature with an experimentally robust training methodology using MIP.

\section{Preliminaries}
\label{sec:pre}

We begin this section by presenting the notation that is used throughout the paper, 
proceed with the definition of the supervised multiclass classification task, and 
conclude with a brief introduction of the notions of data corruption and interpretability 
in machine learning.

\subsection{Notation}
\label{sec:notation}

\begin{itemize}
    \item $F=\{1, \dots, n\}$ is the set of features for positive integer $n\in \mathbb{Z_+}$. 
    \item $C=\{1, \dots, m\}$ is the set of classes for positive integer $m\in \mathbb{Z_+}$.
    \item $I=\{1, \dots, k\}$ and $J=\{k+1, \dots, k+l\}$ are the sets of training and test 
instances for positive integers $k\in \mathbb{Z_+}$ and $l\in \mathbb{Z_+}$, respectively.
    \item Document description (i.e., input) $\vec x_{i}$ is a vector of values for instance 
$i\in I\cup J$ over the set of features $F$ such that $\vec x_{i} = (x_{1,i}, \dots, x_{|F|,i}) 
\in \prod^{|F|}_{f=1} D_f$ where $D_f$ is the domain of input for feature $f\in F$. In this paper, 
we assume the domains of input for all features $f\in F$ are binary (i.e., $D_f = \{0,1\}$ for all 
$f\in F$).
    \item $l_i\in C$ is the label for instance $i\in I\cup J$.
    \item $D=\{d_1,\dots, d_{|I\cup J|}\}$ is the dataset such that 
$d_i = \langle \vec x_{i}, l_i\rangle$ for instance $i\in I\cup J$.
\end{itemize}

\subsection{Multiclass Classification}
\label{sec:multi_class}

Given dataset $D$, (supervised) multiclass classification is the task of constructing a function 
$\mathbb{f}:\{0,1\}^{|F|} \rightarrow C$ using the training dataset $\{d_1,\dots, d_{|I|}\}$ to 
correctly predict the label $l_{j}\in C$ of input $\vec x_{j}$ for test instances $j\in J$ 
such that $\mathbb{f}(\vec x_{j}) = l_{j}$.

Multiclass classification task is commonly solved using models 
that optimize primarily the training accuracy of function $\mathbb{f}$ that 
can be defined as $\frac{|\{i | \mathbb{f}(\vec x_{i}) = l_{i}, i\in I\}|}{|I|}$ and secondarily 
the complexity of function $\mathbb{f}$ that is typically measured by the number of (non-zero) 
parameters of function $\mathbb{f}$. 
Intuitively, maximization of training accuracy aims to learn 
the relationship between input $\vec x_{i}$ and label $l_i\in C$ while the minimization of 
complexity aims to select the simplest function $\mathbb{f}$. 
It has been 
experimentally shown that the combined optimization of these two objectives can learn 
function $\mathbb{f}$ with higher test accuracy (i.e., compared to the optimization of the 
training accuracy by itself) by avoiding overfitting of function $\mathbb{f}$ to the training 
instances~\cite{Bhlmann2011}.

In many settings, however, the training dataset might be corrupted which can lower the test 
accuracy of function $\mathbb{f}$. 
In this paper, we focus on learning accurate function $\mathbb{f}$ that is experimentally 
robust to corrupted training datasets, and briefly cover data corruption in machine learning 
next.

\subsection{Data Corruption in Machine Learning} 
\label{sec:rob_ml}

In the context of machine learning, data corruption refers to modifications to 
input $\vec x_{i}$ and/or label $l_i\in C$ of training instance $i\in I$. 
The nature of such modifications can be due to random noise and/or adversarial 
attacks, which can significantly lower the test accuracy of function 
$\mathbb{f}$~\cite{Han2018}.
In this paper, we focus on learning accurate function $\mathbb{f}$ that is 
experimentally robust to random corruption of labels. 



In machine learning, experimental accuracy and experimental robustness are both 
considered as important objectives to measure the performance of function $\mathbb{f}$, 
and have mainly driven the research to solve challenging tasks, such as image 
recognition~\cite{Krizhevsky2012}, speech recognition~\cite{Deng2013} and natural 
language processing~\cite{Collobert2011}. However, such performance increase 
often came at a price of interpretability of the learned function 
$\mathbb{f}$. Orthogonal to the previously covered performance-related objectives, 
interpretability is concerned with the insights function $\mathbb{f}$ provides 
to its users about relationships that are present in the dataset $I\cup J$ 
or the learned function $\mathbb{f}$~\cite{Murdoch2019}. In this paper, we 
focus on learning an interpretable function $\mathbb{f}$, and briefly cover 
interpretable machine learning next.

\subsection{Interpretable Machine Learning} 
\label{sec:int_ml}

Interpretable machine learning is the extraction of knowledge from function 
$\mathbb{f}$ regarding relationships that are present in the dataset $I\cup J$ 
or learned by the function $\mathbb{f}$~\cite{Murdoch2019}. The knowledge 
extracted by function $\mathbb{f}$ is deemed to be relevant if it provides 
insight for its users about the underlying problem it solves. These insights 
often drive actions and discovery, and can be communicated through visualization, 
natural language, or mathematical equations as we will demonstrate in 
Section~\ref{sec:int_res}. In this paper, the following interpretability 
criteria are applied in the modeling stage of our approach to construct 
function $\mathbb{f}$~\cite{Murdoch2019}:
\begin{enumerate}
    \item Complexity: Interestingly, complexity has a dual role in both improving 
the test accuracy (as previously discussed in Section~\ref{sec:multi_class}) and 
the interpretability of function $\mathbb{f}$. That is, when the learned function 
$\mathbb{f}$ has sufficiently small number of non-zero parameters, the user can 
understand how the learned parameters relate the input $\vec x_{i}$ to the label 
$l_i\in C$ for instance $i\in I\cup J$.
    \item Simulatability: A function $\mathbb{f}$ is considered to be simulatable 
if the user can internalize the entire prediction process of function $\mathbb{f}$ 
(e.g., if the user can predict the label $l_i\in C$ given the input $\vec x_{i}$ for 
function $\mathbb{f}$). Since simulatability requires the entire prediction process 
of function $\mathbb{f}$ to be internalized by the user, the computations required 
to make a prediction must be both simple and small in size.
    \item Modularity: A function $\mathbb{f}$ is considered to be modular if 
the subcomponents of function $\mathbb{f}$ can be interpreted independently.
\end{enumerate}

\section{Training Robust and Interpretable Binarized Regression Models}
\label{sec:regr}

In this section, we first describe our model assumptions and choices, and 
then present two equivalent models based on Mixed-Integer Programming (MIP) 
and Pseudo-Boolean Optimization (PBO) for training interpretable binarized 
regression models to perform the multiclass classification task.

\subsection{Model Assumptions and Choices}
\label{sec:mod_ac}

In this section, we describe our model assumptions and choices to perform 
the multiclass classification task. One fundamental assumption we make in 
this paper is the existence of a function 
$\mathbb{g}: \{0,1\}^{|F|} \rightarrow \mathbb{Z}^{|C|}$ such that 
$\mathbb{f}(\vec x_{i}) = \arg\max \mathbb{g}(\vec x_{i})$ holds for all instances 
$i\in I\cup J$~\cite{Rosenfeld2020}. This assumption has two important benefits 
for effectively modeling the multiclass classification task.

The first benefit is that it allows us to learn function $\mathbb{g}$ as a 
binarized Linear Regression model of the form: 
\begin{align}
&\mathbb{g}(\vec x_{i}) =  \sum_{f\in F} \vec{w}_{f} x_{f,i} + \vec b \quad \forall{i\in I}\label{lr1}
\end{align}
where $\vec w_{f} \in \{-1,0,1\}^{|C|}$ and $\vec b\in \mathbb{Z}^{|C|}$. 
Based on the interpretability criteria that are previously listed in 
Section~\ref{sec:int_ml}, the binarized Linear Regression model can be 
considered one of the most interpretable machine learning models. The 
binarized Linear Regression model has:
\begin{enumerate}
    \item low complexity because it has no 
latent parameters (c.f., Deep Neural Networks~\cite{McCulloch1943}) and can easily 
be regularized,
    \item high simulatability due to the binarization of the domain of the 
weight parameters $\vec{w}_{f}\in \{-1,0,1\}^{|C|}$ which gives each value an 
intuitive semantic meaning (i.e., negative impact: -1, no impact: 0 and positive 
impact: 1) for the prediction of a given class $c\in C$.\footnote{Note 
that while letting the domain of the weight parameter $\vec{w}_{f}$ to be integers 
could improve the test accuracy of function $\mathbb{g}$, it would also decrease 
its interpretability since the user does not have a prior knowledge on the meaning 
of the values of function $\mathbb{g}$. That is, arbitrarily high or low values 
of the weight parameters contradicts with the simulatability criterion as such values 
do not have a natural semantic meaning to the user.} Moreover, it utilizes only 
simple additive features that require at most linear number of addition operations 
(in the size of $|F|$) per class $c\in C$ to make its predictions\footnote{Regularization 
can decrease this value significantly in practice as we experimentally show in 
Table~\ref{tab:exp1_mip_pbo}.}, and
    \item high modularity due to the independence of computations carried for 
each output of function $\mathbb{g}$.
\end{enumerate}

The second benefit is that it allows us to optimize the total margin by which the 
training instances $I$ are correctly and incorrectly classified. We will 
experimentally demonstrate in Section~\ref{sec:exp2} that this optimization allows 
us to learn function $\mathbb{f}$ that is robust to corrupted training datasets.

Next, we present two equivalent models based on Mixed-Integer Programming (MIP) 
and Pseudo-Boolean Optimization (PBO) for training robust and interpretable 
binarized regression models, that are based on our modeling assumptions and 
choices, to perform the multiclass classification task.

\subsection{Mixed-Integer Programming Model}
\label{sec:mip}

In this section, we present the Mixed-Integer Programming (MIP) 
model to train a binarized regression model to perform the multiclass 
classification task.

\paragraph*{Hyperparameters}

The MIP model uses the following hyperparameters:
\begin{itemize}
    \item $\alpha \in \mathbb{R_+}$ is the hyperparameter representing the importance 
of maximizing the total margin of correctly classified training instances.
    \item $\beta \in \mathbb{R_+}$ is the hyperparameter representing the importance 
of minimizing the total margin of incorrectly classified training instances.
\end{itemize}

\paragraph*{Decision Variables}

The MIP model uses the following decision variables:
\begin{itemize}
    \item $w^{+}_{f,c}\in \{0,1\}$ and $w^{-}_{f,c}\in \{0,1\}$ together encode the value of the weight 
between feature $f\in F$ and class $c\in C$.
    \item $b_{c}\in \mathbb{Z}$ encodes the value of the bias for class $c\in C$.
    \item $y_{c,i}\in \mathbb{Z}$ encodes the value of the prediction for class $c\in C$ and training 
instance $i\in I$.
    \item $e^{+}_{i}\in \mathbb{Z_+}$ and $e^{-}_{i}\in \mathbb{Z_+}$ encode the value of the margin for 
(i) correctly and (ii) incorrectly classifying training instance $i\in I$, respectively.
\end{itemize}

\paragraph*{Constraints}

The MIP model uses the following constraints:
\begin{align}
&w^{+}_{f,c} + w^{-}_{f,c} \leq 1 &\forall{f\in F, c \in C}\label{mip1}\\
&\sum_{f\in F}(w^{+}_{f,c} - w^{-}_{f,c}) x_{f,i} + b_{c} = y_{c,i} &\forall{c \in C, i\in I}\label{mip2}\\
&y_{l_i,i} \geq y_{c,i} + e^{+}_{i} - e^{-}_{i} &\forall{i\in I, c \in C \setminus l_i}\label{mip3}\\
&e^{+}_{i}\geq 0, e^{-}_{i} \geq 0 &\forall{i\in I}\label{mip4}\\
&w^{+}_{f,c}, w^{-}_{f,c} \in \{0,1\} &\forall{f\in F, c \in C}\label{mip5}\\
&e^{+}_{i}, e^{-}_{i} \in \mathbb{Z} &\forall{i\in I}\label{mip6}\\
&y_{c,i}\in \mathbb{Z} &\forall{c \in C, i\in I}\label{mip7}\\
&b_{c}\in \mathbb{Z} &\forall{c \in C}\label{mip8}
\end{align}
where constraint (\ref{mip1}) ensures that the weight between feature $f\in F$ and class $c\in C$ 
cannot be both positive and negative, constraint (\ref{mip2}) computes the prediction as the weighted 
sum of its inputs plus the bias for class $c\in C$ and training instance $i\in I$, constraint 
(\ref{mip3}) relates the prediction of all classes $C$ to the margin of correctly and incorrectly 
classifying training instance $i\in I$, constraint (\ref{mip4}) enforces the margin of correctly 
and incorrectly classifying training instance $i\in I$ to be non-negative and constraints 
(\ref{mip5}-\ref{mip8}) define the domain of all decision variables.

\paragraph*{Objective Function}

The MIP model uses the following objective function:
\begin{align}
&\min \; -\alpha \sum_{i \in I} e^{+}_{i} + \beta \sum_{i \in I} e^{-}_{i} + 
\sum_{f\in F, c\in C}(w^{+}_{f,c} + w^{-}_{f,c}) \label{mip0}
\end{align}
which balances the optimization of prediction margin and model size by (i) maximizing the total 
margin of correctly classified training instances, (ii) minimizing the total margin of incorrectly 
classified training instances and (iii) minimizing the total model size.


\subsection{Pseudo-Boolean Optimization Model}
\label{sec:pbo}

In this section, we present the Pseudo-Boolean Optimization (PBO) model to train a binarized 
regression model to perform the multiclass classification task.

\paragraph*{Hyperparameters}

The PBO model uses the same sets of hyperparameters as the previously presented MIP model 
in addition to hyperparameter $Q$ that 
is used to quantize the domains of integer-valued decision variables.

\paragraph*{Decision Variables, Constraints and Objective Function}

The PBO model uses the same sets of decision variables as the previously presented MIP model 
to encode the weights $w^{+}_{f,c}\in \{0,1\}$ and 
$w^{-}_{f,c}\in \{0,1\}$. Given PBO is restricted to have decision variables with only 
binary domains, the PBO model uses a set of decision variables $b_{c,q}\in \{0,1\}$, 
$y_{c,i,q}\in \{0,1\}$, $e^{+}_{i,q}\in \{0,1\}$ and $e^{-}_{i,q}\in \{0,1\}$ to equally
represent the domains of decision variables 
$b_{c}\in \mathbb{Z}$, $y_{c,i}\in \mathbb{Z}$, $e^{+}_{i}\in \mathbb{Z_+}$ and 
$e^{-}_{i}\in \mathbb{Z_+}$, respectively. The domain of some (bounded) decision 
variable $x\in \mathbb{Z}$ is represented using the following formula\footnote{While we do not work with 
real-valued decision variables in this paper, it is important to note that a similar quantization
formula can be used to approximate the domains of real-valued decision variables~\cite{Say2020}.}:
\begin{align}
&x = x_{LB} + \sum_{q=1}^{Q} 2^{q-1}x_{q}\label{expr_formula}
\end{align}
together with the following constraint:
\begin{align}
&x_{LB} + \sum_{q=1}^{Q} 2^{q-1}x_{q} \leq x_{UB} \label{quantization_constraint}
\end{align}
where $x_{LB}$ and $x_{UB}$ represent the lower bound and the upper bound on decision 
variable $x$, and the value of hyperparameter $Q$ is calculated according to the following 
formula:
\begin{align}
&Q = \left \lceil{\log_2 (x_{UB} - x_{LB} + 1)}\right\rceil \label{Q_formula}
\end{align}



Finally, the PBO model uses the same sets of constraints 
and the objective function as the previously presented MIP model 
with the minor modification that encodes the set of decision 
variables $b_{c}\in \mathbb{Z}$, $y_{c,i}\in \mathbb{Z}$, 
$e^{+}_{i}\in \mathbb{Z_+}$ and $e^{-}_{i}\in \mathbb{Z_+}$ 
according to formula (\ref{expr_formula}), formula (\ref{Q_formula}) 
and constraint (\ref{quantization_constraint}) using the set of 
decision variables $b_{c,q}\in \{0,1\}$, $y_{c,i,q}\in \{0,1\}$, 
$e^{+}_{i,q}\in \{0,1\}$ and $e^{-}_{i,q}\in \{0,1\}$.





\section{Experimental Results}
\label{sec:exp}

In this section, we begin by presenting the results of two sets of 
computational experiments. In the first set of experiments, we test 
the effectiveness of training a binarized regression model to perform 
the multiclass classification task using our MIP and PBO models against
Logistic Regression (LR) with real-valued weights and biases, and Gradient 
Descent (GD) with binarized weights and integer-valued biases over 
three standard datasets, namely: \textit{MNIST}~\cite{Lecun2010}, 
\textit{Flags}~\cite{Romano2021} and \textit{Ask Ubuntu}~\cite{Braun2017}. 
We provide detailed comparative results on 
both the training and test accuracy of all models, and further report 
the model sizes, runtimes and optimality gaps of our MIP and PBO models. 
Our detailed results for the first set of experiments show that the MIP 
model achieves competitive test accuracy results to the real-valued 
LR model over three datasets and outperforms the LR model in one dataset, 
while achieving upto around 70\% model size reduction. In the second set of 
experiments, we test the effectiveness of training a binarized regression 
model to perform the multiclass classification task using our MIP model 
against the previously described LR and GD models over three corrupted 
datasets. We provide detailed comparative results on the test accuracy 
of all models. 
Our results for the second set of experiments show that our MIP model 
outperforms the LR model in majority of the corrupted datasets without
experiencing significant degradation in test accuracy. We conclude this 
section by visually demonstrating the interpretability of our MIP model 
using the interpretability criteria discussed previously in 
Section~\ref{sec:int_ml}.

\subsection{Experimental Domains and Setup}
\label{sec:exp_set}

The domains and the setup used throughout this section is as follows.

\paragraph*{Experimental Domains}

The MNIST~\cite{Lecun2010} dataset consists of 784 features and 10 
classes with 70,000 instances. The dataset is of grayscale images 
of handwritten digits represented by 784 pixels (i.e., in 28 by 28 pixel 
format) where each pixel has a value between 0 and 255. We have binarized 
each pixel such that the pixel is set to 1 if its value is greater than 
$\frac{255}{2}$, and 0 otherwise. We trained 
our models over 20, 60, 100, 200, 300, 500, 700 and 1000 training 
instances, and reported test accuracy results over the remaining 
dataset. The Flags~\cite{Romano2021} dataset consists of 43 features 
and 5 classes with 143 instances. The dataset describes the attributes 
of the flags of various countries. We trained our models over 10, 25, 
35 and 50 training instances, and reported test accuracy results over 
the remaining dataset. Finally, the Ask Ubuntu~\cite{Braun2017} dataset 
consists of 5 classes with 162 instances. The dataset has five classes 
which refer to the user's intent behind the questions asked. Natural 
language preprocessing is performed to extract the features of this 
dataset. We trained our models over 10, 23 and 43 training 
instances, and reported test accuracy results over the remaining 
dataset. Finally in the second set of experiments, 10\% of labels are 
randomly assigned to a different class, and everything else remained 
the same. Table~\ref{tab:dom} provides a summary of the datasets.


\begin{table}
  \centering
    \begin{tabular}{| l | l | l | l | l|}
    \hline
    Dataset & $|$F$|$, $|$C$|$ & $|$I$|$ & $\alpha$, $\beta$ & Brief Description \\ \hline
    MNIST & 784, 10 & 20, 60, & 5, 10 & Image classification task \\
    & & 100, 200, & & with grayscale images of \\ 
    & & 300, 500,  & & handwritten digits \\ 
    & & 700, 1000 & & represented by 784 pixels \\
    & & & & (i.e., in 28 by 28 format). \\ \hline
    Flags & 43, 5 & 10, 25, & 2, 5 & Describes the attributes \\
    & & 35, 50 & & of the flags of countries. \\ \hline
    Ubuntu & 48 & 10 & 2, 5 & Natural language processing \\
    & 93 & 23 & & task to classify the intention \\ 
    & 153 & 43 & & of the user behind the \\
    &  &  & & question asked. \\ \hline
  \end{tabular}
  \caption{Summary of the datasets including feature and class sizes, number of 
training instances used for each training problem, the values of the 
hyperparameters $\alpha$ and $\beta$, and the brief descriptions of the datasets.}
  \label{tab:dom}
\end{table}

\paragraph*{Experimental Setup}

All experiments were run on Intel Core i5-8250U CPU 1.60GHz with 8.00 GB 
memory, using a single thread with one hour total time limit per training 
problem. The MIP model is optimized using Gurobi~\cite{Gurobi2021}, the 
PBO model is optimized using RoundingSat~\cite{Elffers2018}, the LR model is 
optimized using Scikit-learn~\cite{Pedregosa2011} and the GD model is created 
using LARQ~\cite{Geiger2020} and trained with a learning rate of 0.001 over 
200 epochs using Adam~\cite{Kingma2014}. The value of hyperparameter $\alpha$ 
is selected using grid search where the value of hyperparameter $\beta$ 
is set to $2\alpha$. Both MIP and PBO models used the same selected values 
of hyperparameters $\alpha$ and $\beta$ as detailed in Table~\ref{tab:dom}. 

\subsection{Experiment 1: Training with Standard Datasets}
\label{sec:exp1}

\begin{figure}
\centering
    \begin{subfigure}{.24\textwidth}
    \centering
        \includegraphics[width=\linewidth]{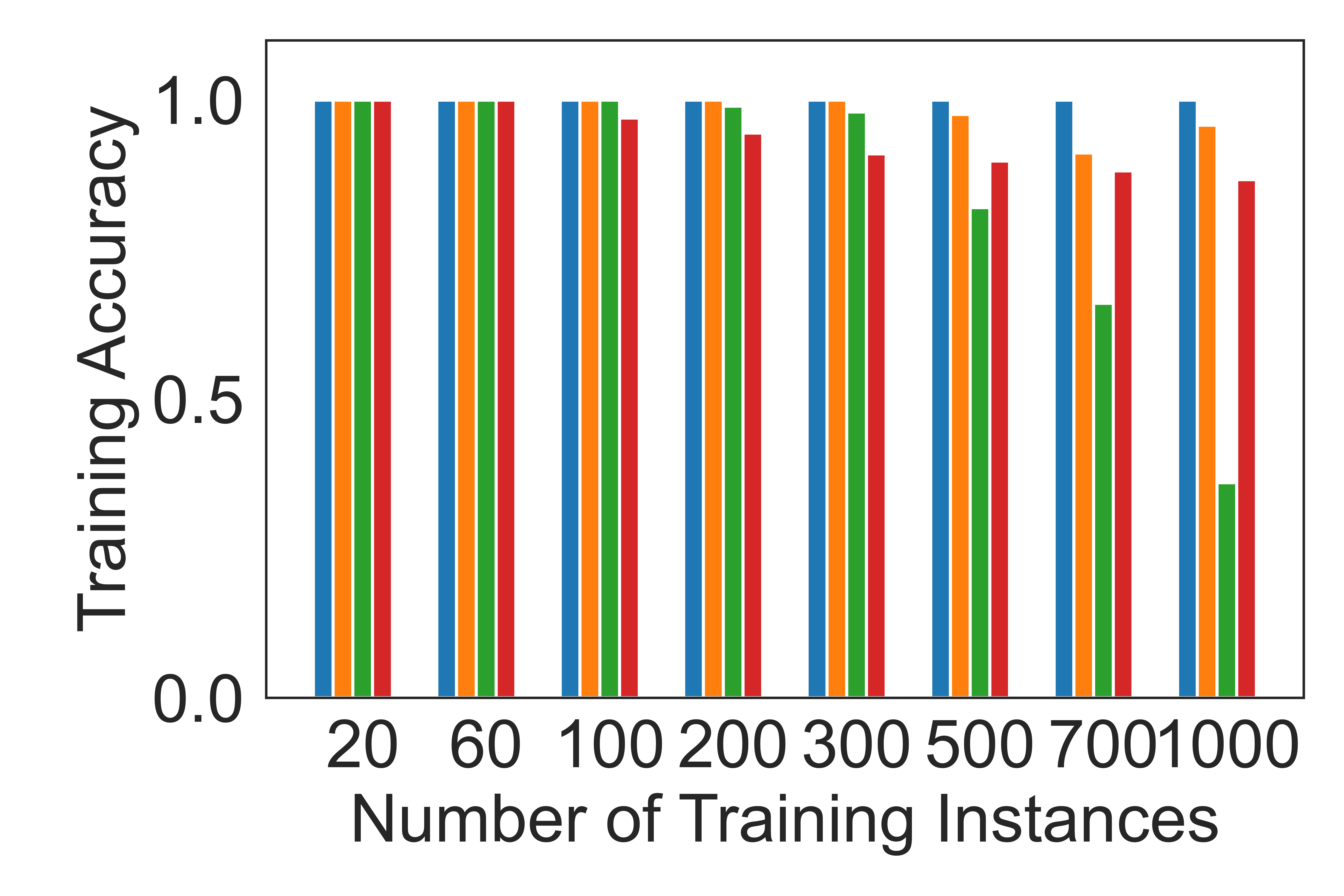}
        \caption{Standard MNIST}
        \label{fig:train_mnist}
    \end{subfigure}
    \begin{subfigure}{.24\textwidth}
    \centering
        \includegraphics[width=\linewidth]{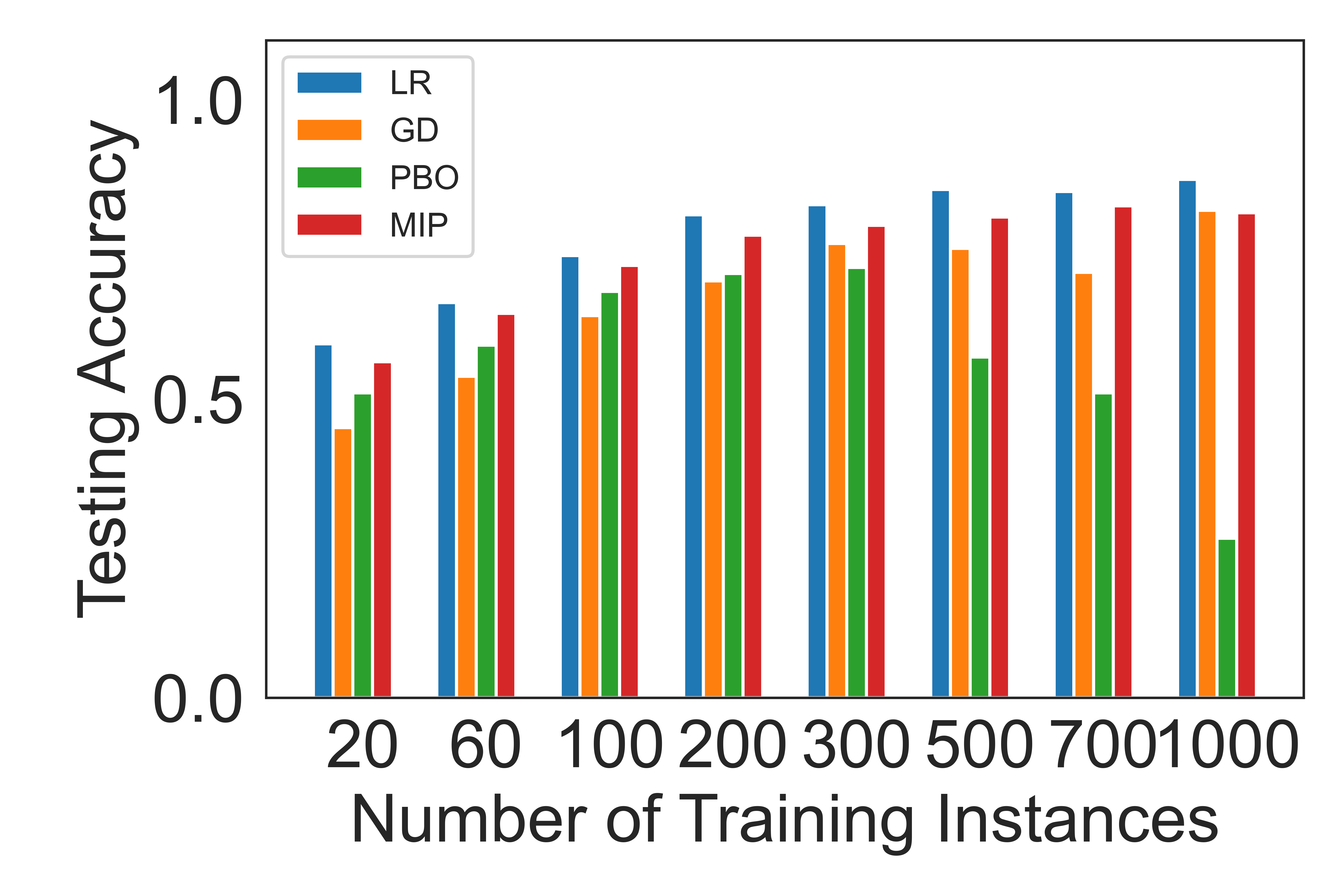}
        \caption{Standard MNIST}
        \label{fig:test_mnist}
    \end{subfigure}
    
    \begin{subfigure}{.24\textwidth}
    \centering
        \includegraphics[width=\linewidth]{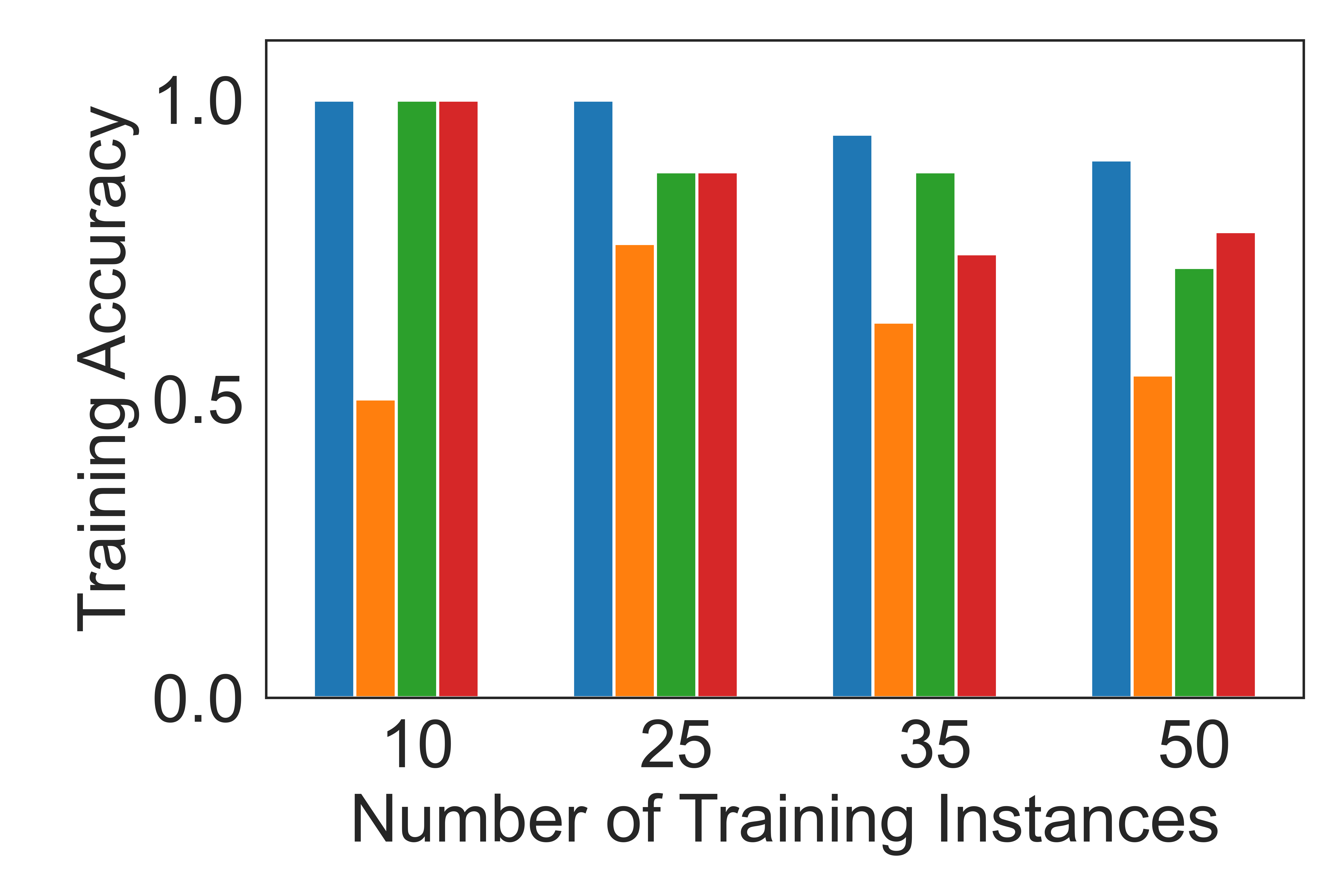}
        \caption{Standard Flags}
        \label{fig:train_flags}
    \end{subfigure}
    \begin{subfigure}{.24\textwidth}
    \centering
        \includegraphics[width=\linewidth]{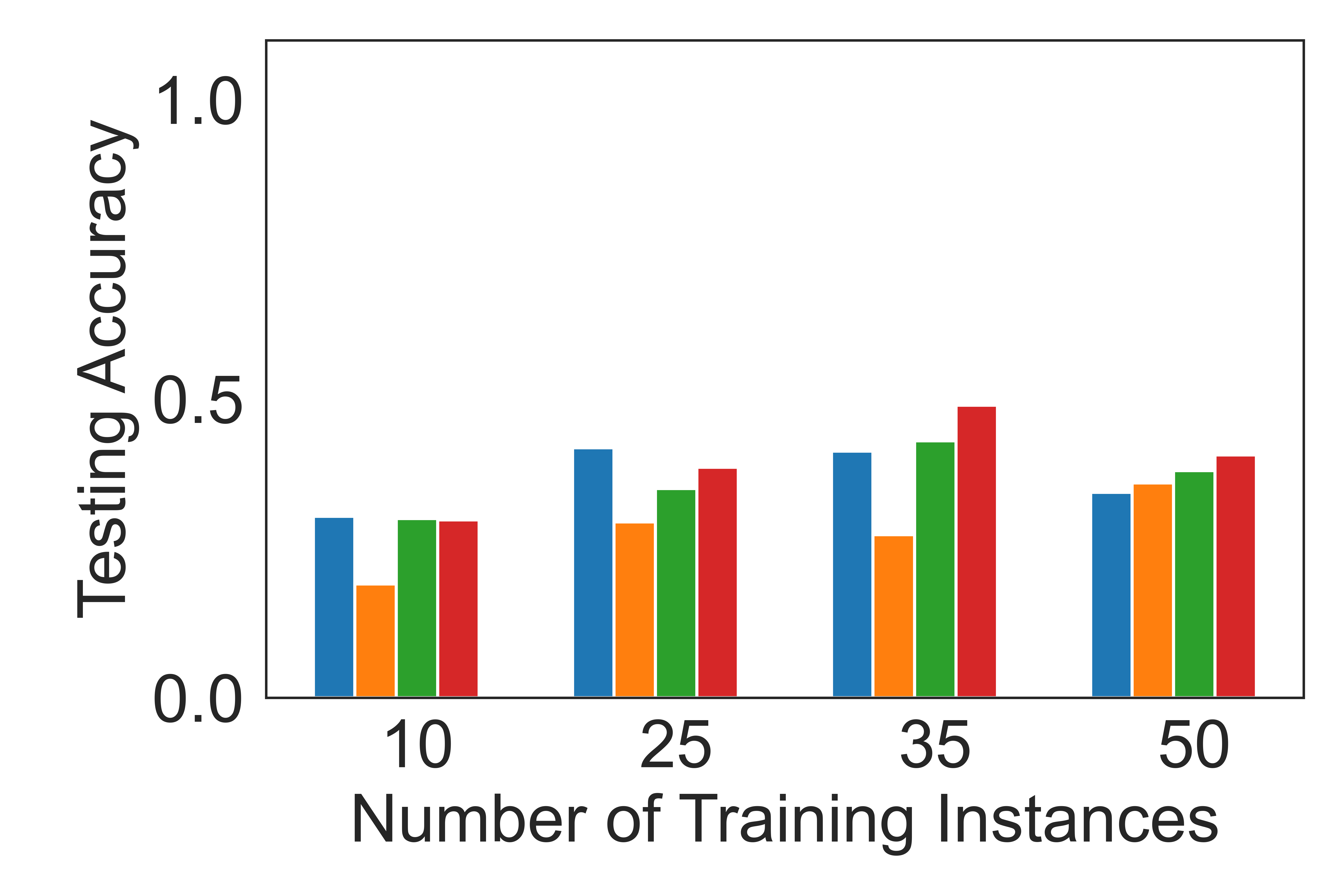}
        \caption{Standard Flags}
        \label{fig:test_flags}
    \end{subfigure}
    
    \begin{subfigure}{.24\textwidth}
    \centering
        \includegraphics[width=\linewidth]{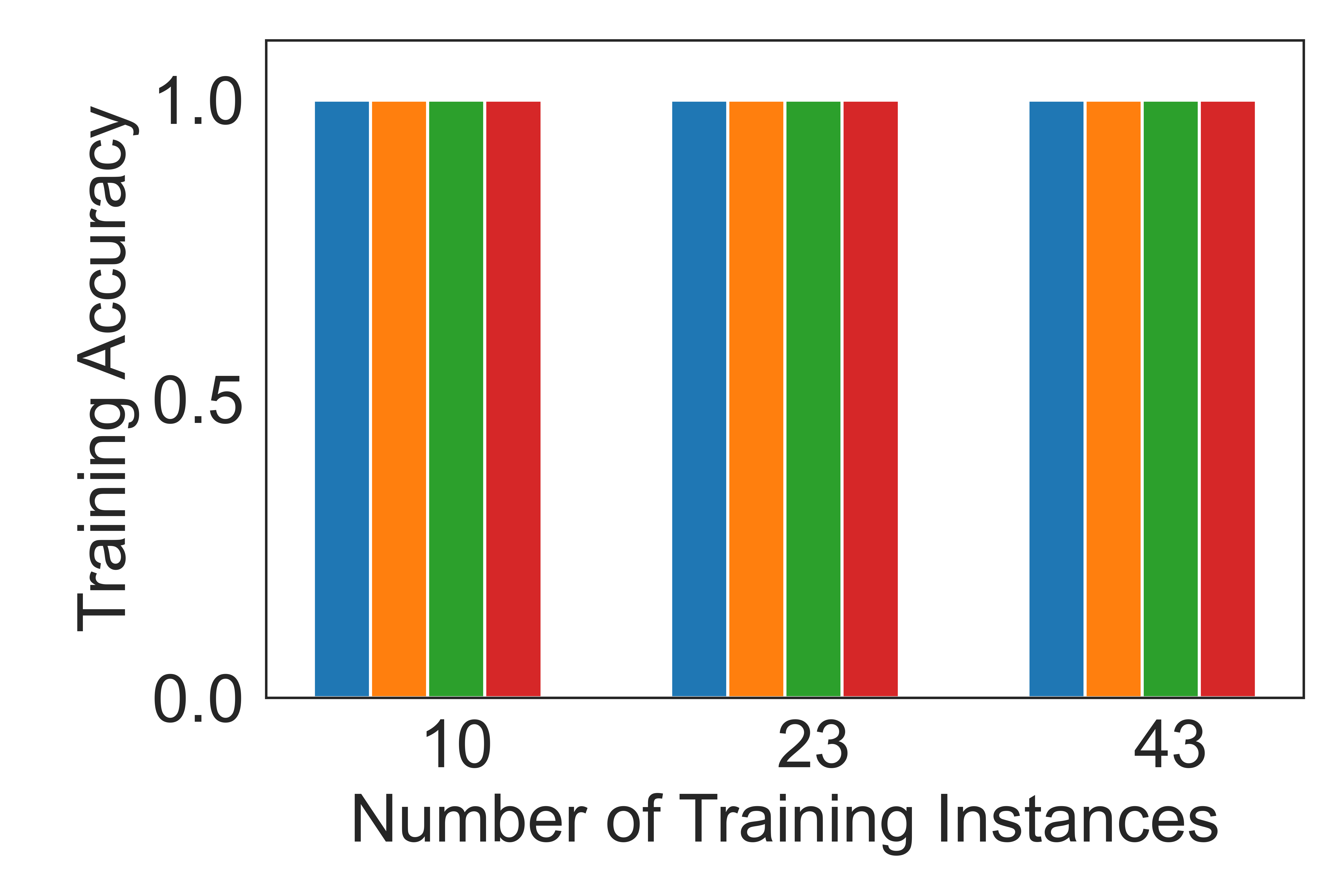}
        \caption{Standard Ask Ubuntu}
        \label{fig:train_ubuntu}
    \end{subfigure}
    \begin{subfigure}{.24\textwidth}
    \centering
        \includegraphics[width=\linewidth]{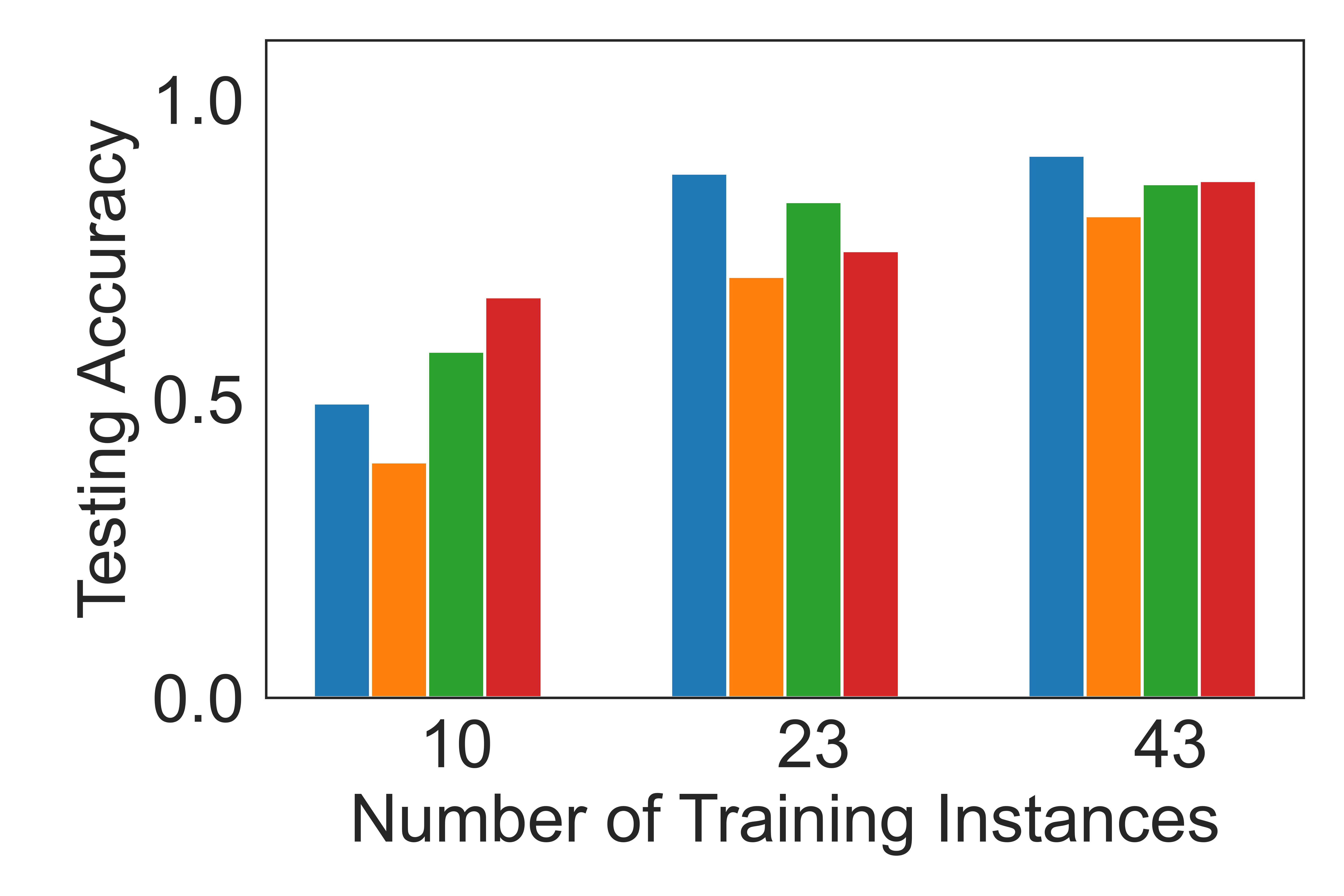}
        \caption{Standard Ask Ubuntu}
        \label{fig:test_ubuntu}
    \end{subfigure}
    \caption{Visualization of Experiment 1 which compares the training 
accuracy (left) and the test accuracy (right) of the MIP model 
(red) and the PBO model (green) against the LR model (blue) and the GD 
model (orange) over three standard datasets and different number of 
training instances within one hour time limit.}
  \label{fig:exp1}
\end{figure}

In the first set of experiments, we compare the effectiveness of performing 
the multiclass classification task using the MIP model and the PBO model 
against LR and GD over the standard datasets. 
Figure~\ref{fig:exp1} visualizes the training 
accuracy (i.e., the left column) and the test accuracy (i.e., the right column) 
of all models over each dataset (i.e., represented by individual rows). 
Table~\ref{tab:exp1_mip_pbo} compares the model sizes, runtimes and optimality 
gaps of both MIP and PBO models.

In Figure~\ref{fig:exp1}, the inspection of subfigures 
\ref{fig:train_mnist} and \ref{fig:test_mnist} highlights the clear 
capability of the MIP model to scale with the increasing size of 
the training datasets. In the MNIST dataset, the MIP model 
outperforms the binarized GD model in 7 out of 8 training problems 
in terms of testing accuracy, and performs competitively with the 
real-valued LR model. In contrast to the MIP model, the performance 
of the PBO model degrades with the increasing number of training instances. 
Moreover, the inspection of subfigures \ref{fig:train_flags}, 
\ref{fig:test_flags}, \ref{fig:train_ubuntu} and \ref{fig:test_ubuntu} 
demonstrates the ability of both the MIP model and the PBO model to train 
models with competitive test accuracies over small training problems. In 
Flags and Ask Ubuntu datasets, the MIP model and the PBO model consistently 
outperform the binarized GD model in terms of testing accuracy, and perform 
competitively with the real-valued LR model. 

\begin{table}
  \centering
    \begin{tabular}{| l | l | l | l | l|}
    \hline
    Model & Dataset, $|$I$|$ & Gap & Time (sec.) & Reduction (\%) \\ \hline
    \textbf{MIP} & MNIST, 20 & \textbf{0.00} & 3.99 & 62.19 \\ \hline
    PBO & MNIST, 20 & 0.34 & x $\geq$ T.O. & 3.48 \\ \hline
    \textbf{MIP} & MNIST, 60 & \textbf{0.00} & 304.87 & 55.23 \\ \hline
    PBO & MNIST, 60 & 0.52 & T.O. & 11.28 \\ \hline
    \textbf{MIP} & MNIST, 100 & \textbf{0.00034} & T.O. & 49.72 \\ \hline
    PBO & MNIST, 100 & 0.51 & T.O. & 17.48 \\ \hline
    \textbf{MIP} & MNIST, 200 & \textbf{0.00063} & T.O. & 44.87 \\ \hline
    PBO & MNIST, 200 & 0.85 & T.O. & 34.22 \\ \hline
    \textbf{MIP} & MNIST, 300 & \textbf{0.00068} & T.O. & 43.52 \\ \hline
    PB & MNIST, 300 & 1.28 & T.O. & 31.98 \\ \hline
    \textbf{MIP} & MNIST, 500 & \textbf{0.00078} & T.O. & 40.47 \\ \hline
    PBO & MNIST, 500 & 9.76 & T.O. & 19.97 \\ \hline
    \textbf{MIP} & MNIST, 700 & \textbf{0.0010} & T.O. & 39.31 \\ \hline
    PBO & MNIST, 700 & 12.60 & T.O. & 15.30 \\ \hline
    \textbf{MIP} & MNIST, 1000 & \textbf{0.00081} & T.O. & 37.33 \\ \hline
    PBO & MNIST, 1000 & 140.89 & T.O. & 6.54 \\ \hline
    \textbf{MIP} & Flags, 10 & \textbf{0.00} & 0.13 & 69.77 \\ \hline
    PBO & Flags, 10 & 0.00 & 22.00 & 53.48 \\ \hline
    \textbf{MIP} & Flags, 25 & \textbf{0.00} & 0.71 & 48.37 \\ \hline
    PBO & Flags, 25 & 96.25 & T.O. & 51.62 \\ \hline
    \textbf{MIP} & Flags, 35 & \textbf{0.00} & 0.58 & 41.40 \\ \hline
    PBO & Flags, 35 & 72.03 & T.O. & 47.44 \\ \hline
    \textbf{MIP} & Flags, 50 & \textbf{0.00} & 2.43 & 44.19 \\ \hline
    PB & Flags, 50 & 119.56 & T.O. & 37.67 \\ \hline
    \textbf{MIP} & Ubuntu, 10 & \textbf{0.00} & 0.11 & 69.58 \\ \hline
    PBO & Ubuntu, 10 & 0.00 & 13.16 & 28.75 \\ \hline
    \textbf{MIP} & Ubuntu, 23 & \textbf{0.00} & 0.35 & 68.17 \\ \hline
    PBO & Ubuntu, 23 & 0.00 & 1192.20 & 33.54 \\ \hline
    \textbf{MIP} & Ubuntu, 43 & \textbf{0.00} & 0.83 & 65.88 \\ \hline
    PBO & Ubuntu, 43 & 251.08 & T.O. & 31.50 \\ \hline
  \end{tabular}
  \caption{Comparison of the MIP model and the PBO model in terms of their 
duality gaps, runtimes and model size reductions in Experiment 1 where T.O. 
means timeout.}
  \label{tab:exp1_mip_pbo}
\end{table}

The inspection of Table~\ref{tab:exp1_mip_pbo} highlights the benefit of 
using the MIP model over the PBO model for training binarized regression models. 
Across all training problems, the MIP model outperforms the PBO 
model in terms of runtime performance where the PBO model almost always runs out 
of the one hour time limit. As a result, the PBO model can return feasible 
solutions with large duality gaps that can yield low testing accuracy 
(e.g., as visualized in subfigure~\ref{fig:test_mnist}). 


\subsection{Experiment 2: Training with Corrupted Datasets}
\label{sec:exp2}

\begin{figure}
\centering
    \begin{subfigure}{.24\textwidth}
    \centering
        \includegraphics[width=\linewidth]{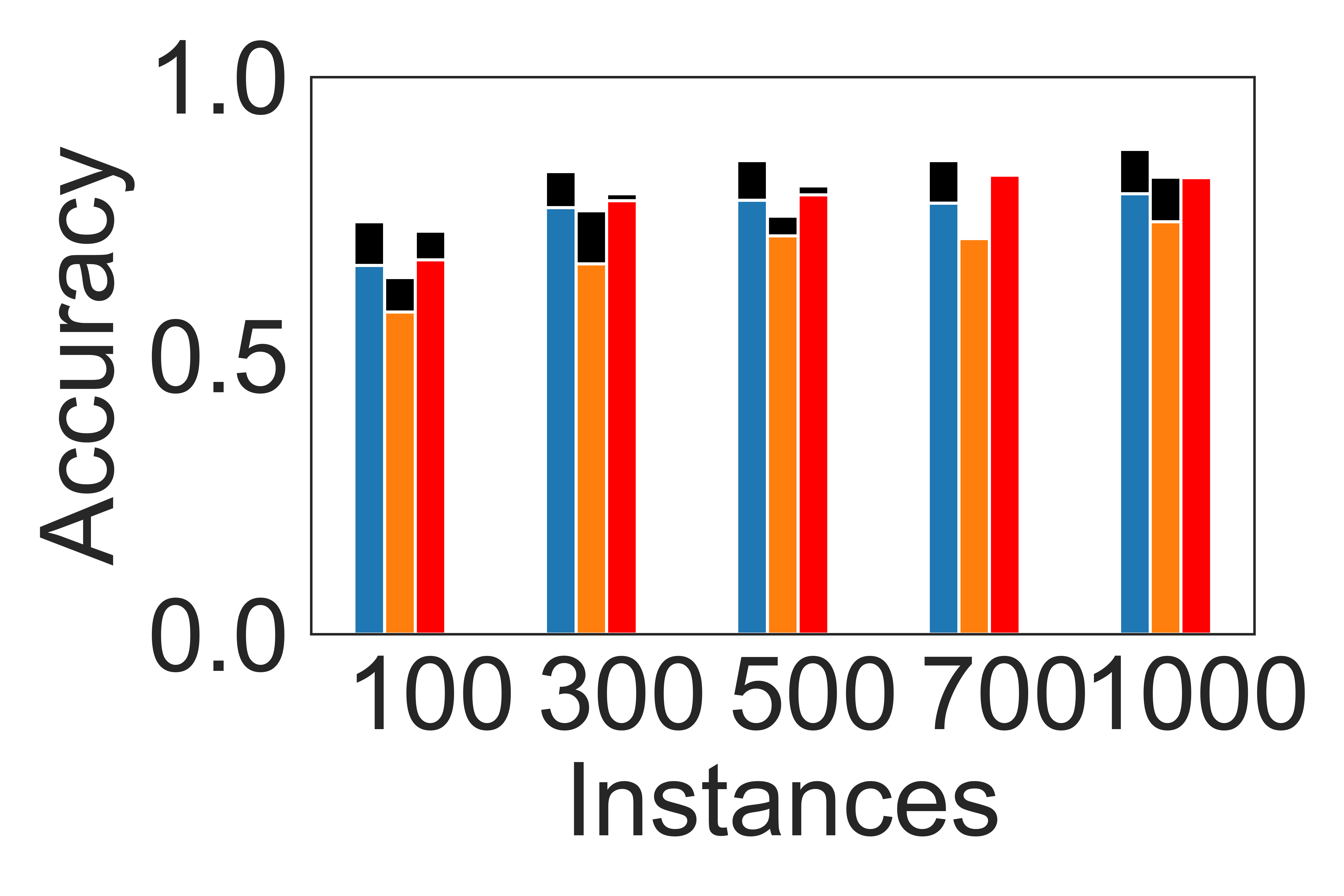}
        \caption{Corrupted MNIST}
        \label{fig:corr_mnist}
    \end{subfigure}
    
    \begin{subfigure}{.24\textwidth}
    \centering
        \includegraphics[width=\linewidth]{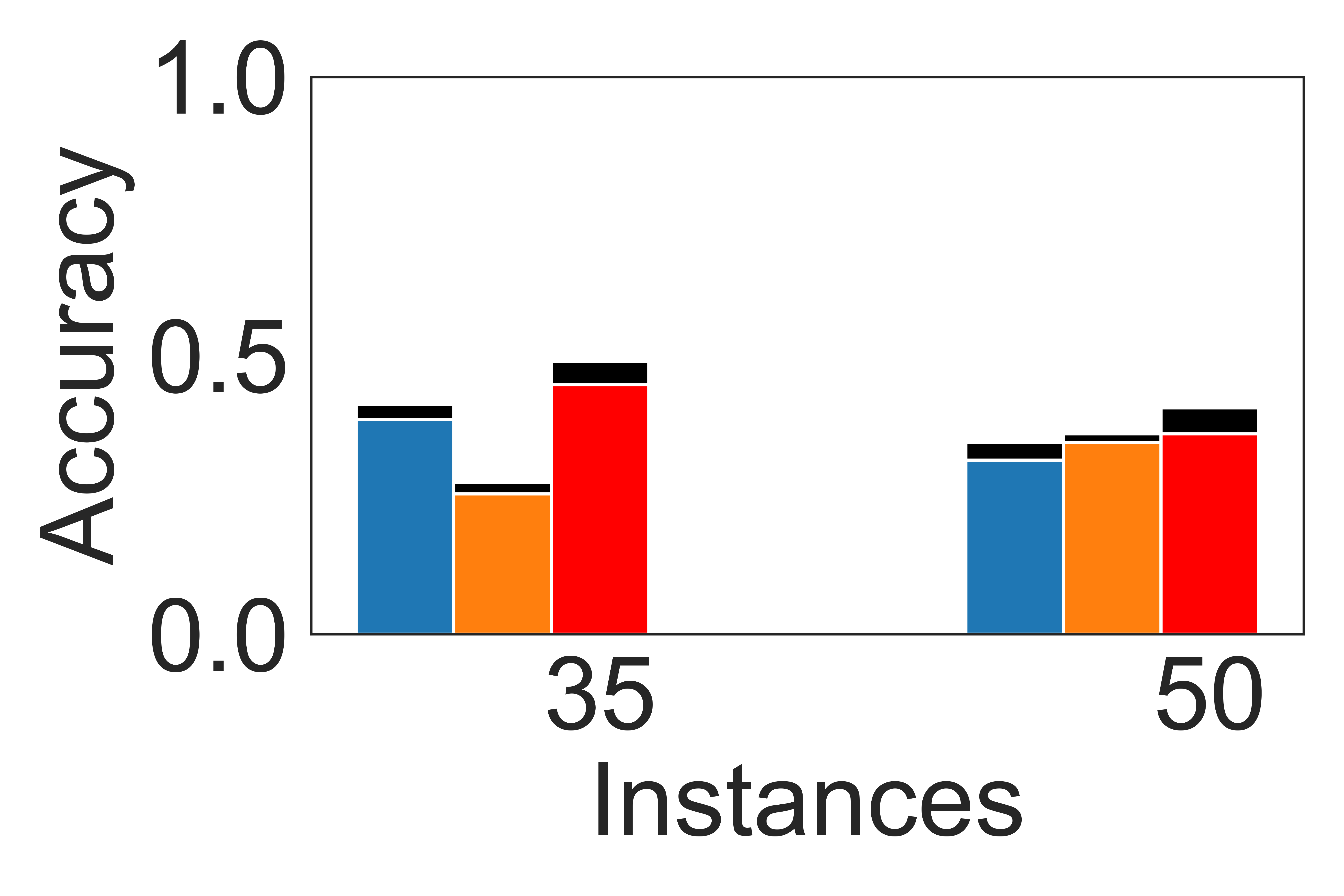}
        \caption{Corrupted Flags}
        \label{fig:corr_flags}
    \end{subfigure}
    
    \begin{subfigure}{.24\textwidth}
    \centering
        \includegraphics[width=\linewidth]{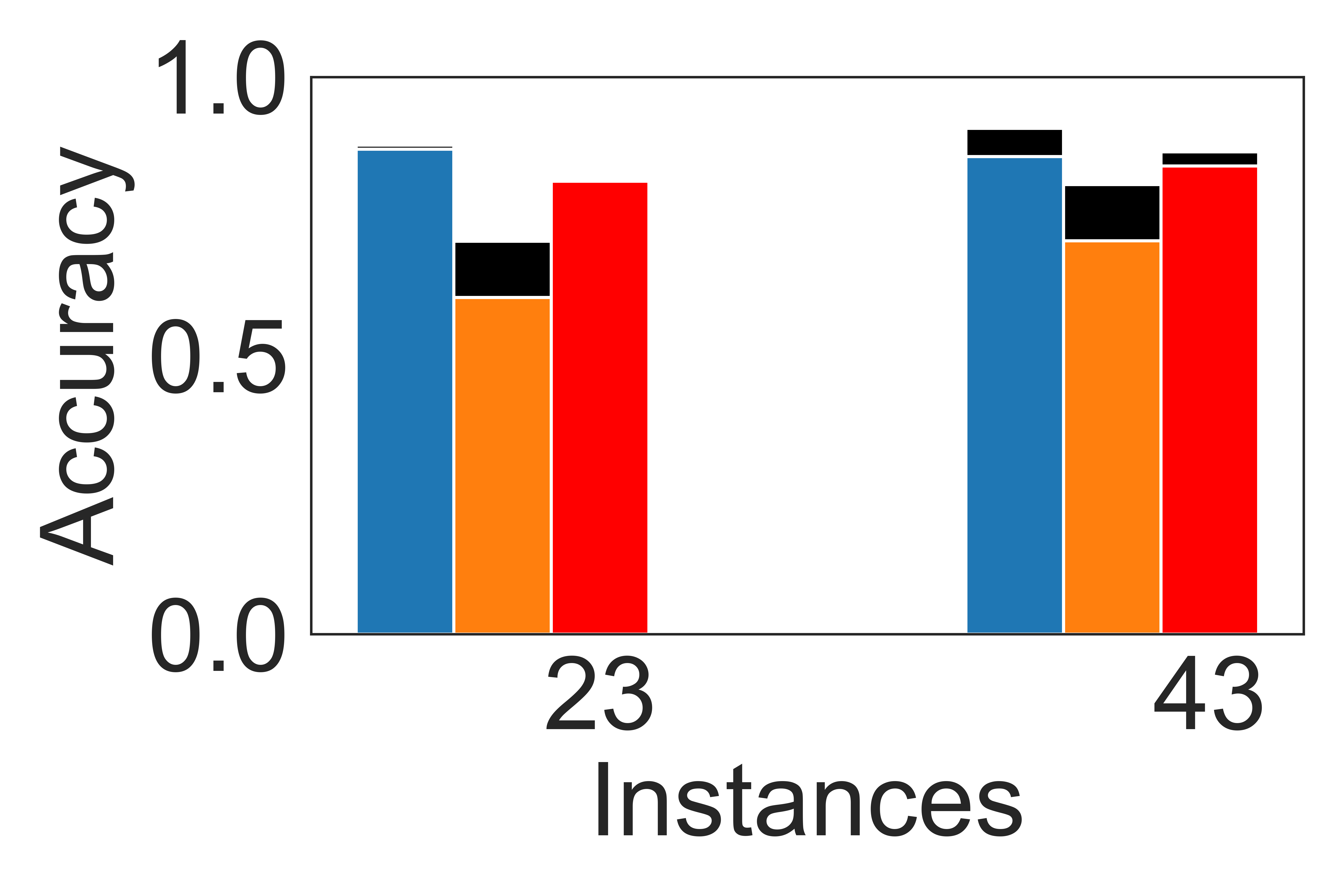}
        \caption{Corrupted Ask Ubuntu}
        \label{fig:corr_ubuntu}
    \end{subfigure}
    \caption{Visualization of Experiment 2 which compares the test accuracy of 
the MIP model (red) against the LR model (blue) and the GD model (orange) over 
three corrupted datasets and different number of training instances 
within one hour time limit. For each model, the black region represents the 
decrease in test accuracy due to random corruption of labels. 
}
  \label{fig:exp2}
\end{figure}

In the second set of experiments, we compare the effectiveness of performing 
the multiclass classification task using the MIP model against LR and GD models 
over corrupted datasets. Figure~\ref{fig:exp2} visualizes the test accuracy 
of all three models over each corrupted dataset (i.e., represented by individual 
rows) as well as the decrease in test accuracy that is due to data corruption, 
which is measured by the difference between the test accuracy of a given model 
in Experiment 1 and Experiment 2.

In Figure~\ref{fig:exp2}, the inspection of subfigures~\ref{fig:corr_mnist}, 
\ref{fig:corr_flags} and \ref{fig:corr_ubuntu} demonstrates the experimental 
robustness of the MIP model over both LR and GD models across all three 
corrupted datasets. In the corrupted MNIST and Flags datasets, the MIP model 
outperforms both LR and GD models with minimal decrease in its testing 
accuracy. Similarly in the corrupted Ask Ubuntu dataset, the MIP model 
outperforms the GD model while performing competitively with the LR model. 

Overall, our results highlight the clear performance benefit of training 
binarized regression models for the multiclass classification task using 
MIP. Next, we turn our attention to the interpretability our MIP model.

\subsection{Interpretability Results}
\label{sec:int_res}

In this section, we focus on the interpretability of our MIP model based 
on the criteria that are previously discussed in 
Section~\ref{sec:int_ml}, namely: complexity, simulatability and modularity. 
Given the binarized regression model is highly modular by definition (i.e., 
each output of the regression model can be computed independently from the 
others), we focus on the remaining two criteria.

\paragraph*{Complexity}

As evident from Table~\ref{tab:exp1_mip_pbo}, our MIP 
model often learns binarized regression models with significantly reduced 
complexity through near optimal regularization; by setting both decision 
variables $w^{+}_{f,c}\in \{0,1\}$ and $w^{-}_{f,c}\in \{0,1\}$ equal to 0. 
Given the learned binarized regression model is often optimally regularized 
and has no latent parameters by definition, we conclude that the binarized 
regression model learned by solving our MIP model has low complexity.

\paragraph*{Simulatability}

\begin{figure}
\centering
    \begin{subfigure}{.19\textwidth}
    \centering
        \includegraphics[width=\linewidth]{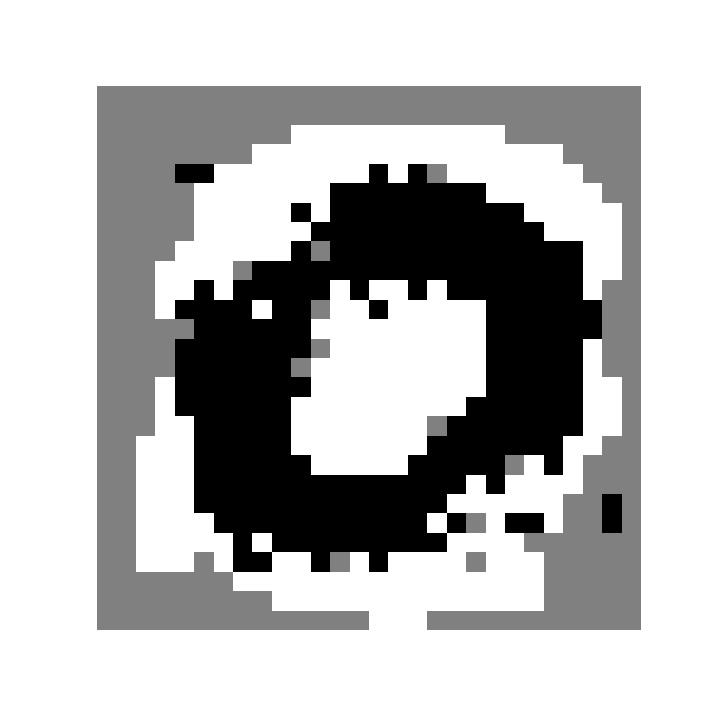}
        \caption{Number 0}
        \label{fig:image_0}
    \end{subfigure}
    \begin{subfigure}{.19\textwidth}
    \centering
        \includegraphics[width=\linewidth]{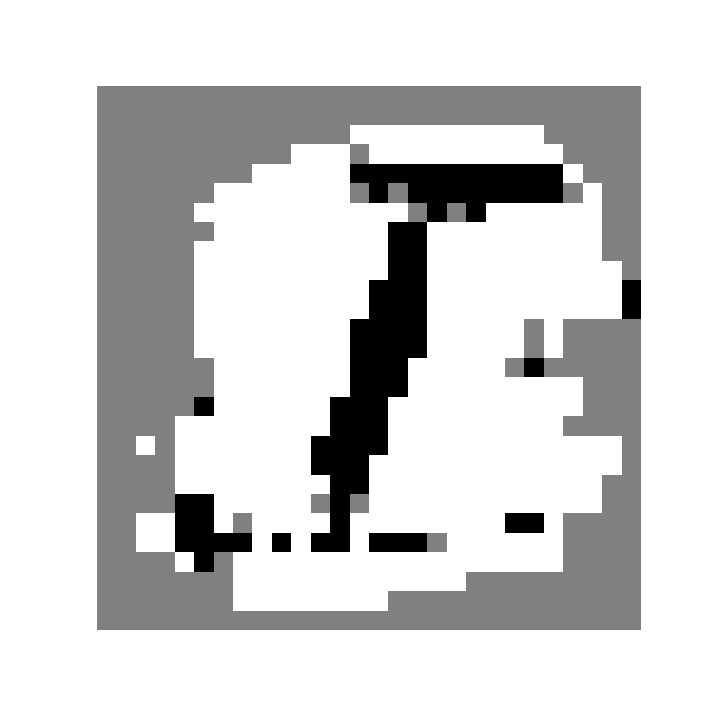}
        \caption{Number 1}
        \label{fig:image_1}
    \end{subfigure}
    
    \begin{subfigure}{.19\textwidth}
    \centering
        \includegraphics[width=\linewidth]{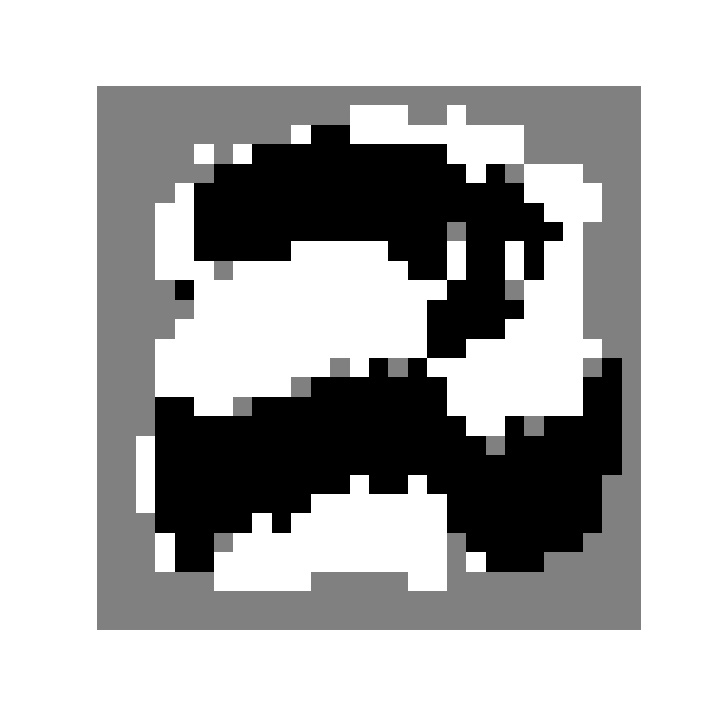}
        \caption{Number 2}
        \label{fig:image_2}
    \end{subfigure}
    \begin{subfigure}{.19\textwidth}
    \centering
        \includegraphics[width=\linewidth]{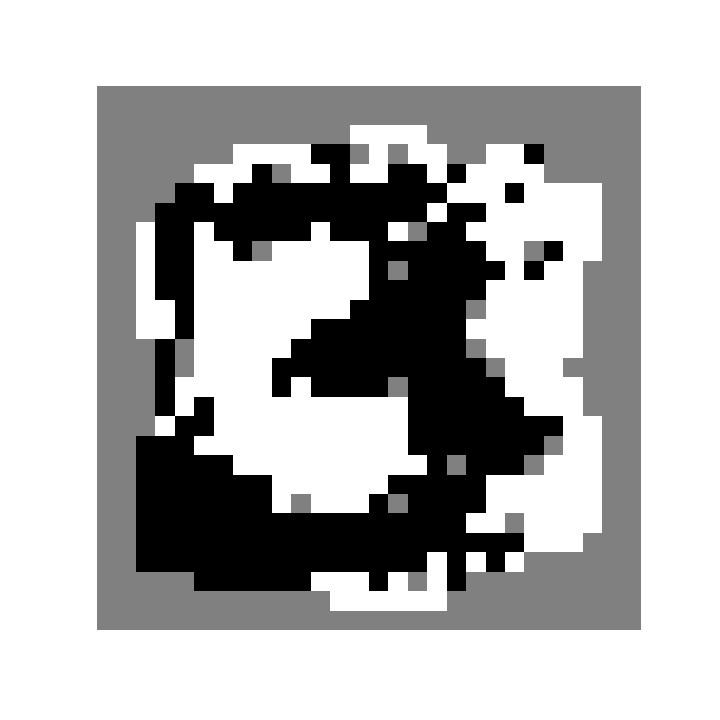}
        \caption{Number 3}
        \label{fig:image_3}
    \end{subfigure}
    
    \begin{subfigure}{.19\textwidth}
    \centering
        \includegraphics[width=\linewidth]{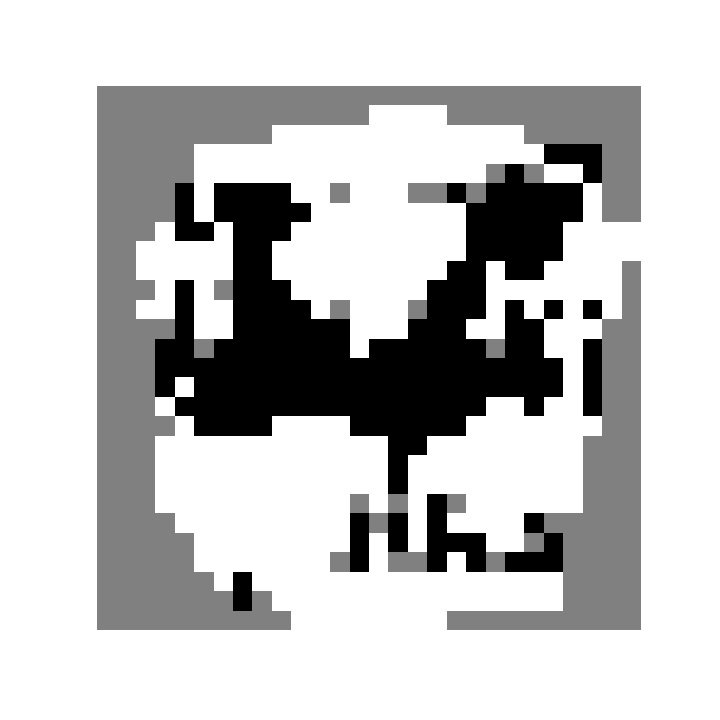}
        \caption{Number 4}
        \label{fig:image_4}
    \end{subfigure}
    \begin{subfigure}{.19\textwidth}
    \centering
        \includegraphics[width=\linewidth]{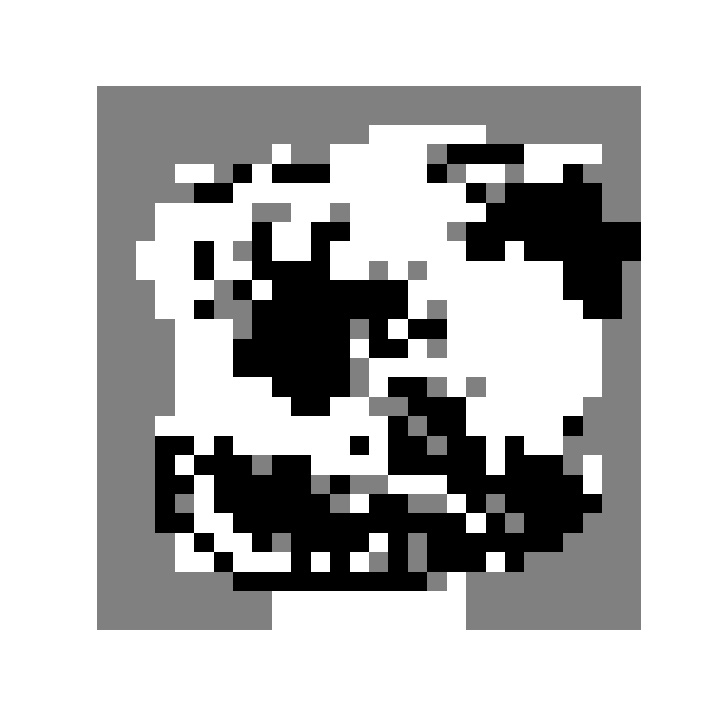}
        \caption{Number 5}
        \label{fig:image_5}
    \end{subfigure}
    
    \begin{subfigure}{.19\textwidth}
    \centering
        \includegraphics[width=\linewidth]{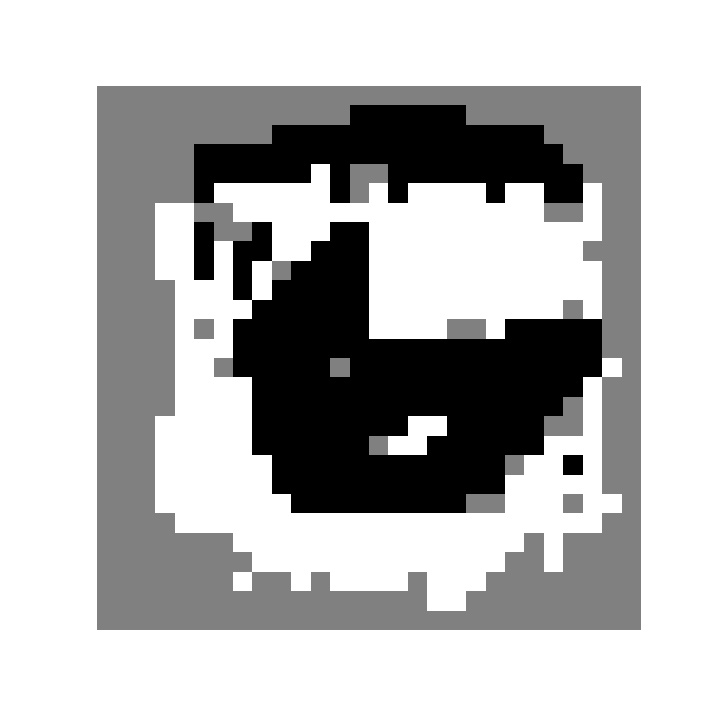}
        \caption{Number 6}
        \label{fig:image_6}
    \end{subfigure}
    \begin{subfigure}{.19\textwidth}
    \centering
        \includegraphics[width=\linewidth]{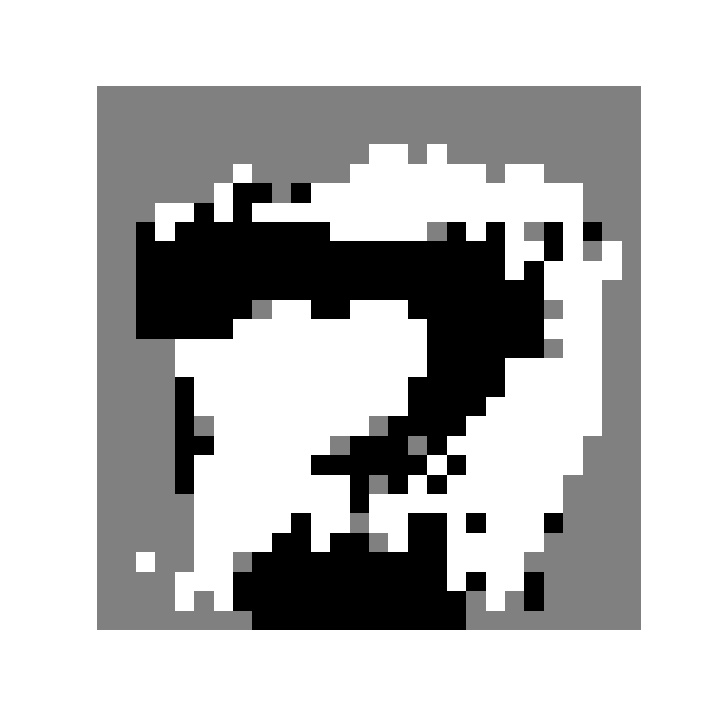}
        \caption{Number 7}
        \label{fig:image_7}
    \end{subfigure}
    
    \begin{subfigure}{.19\textwidth}
    \centering
        \includegraphics[width=\linewidth]{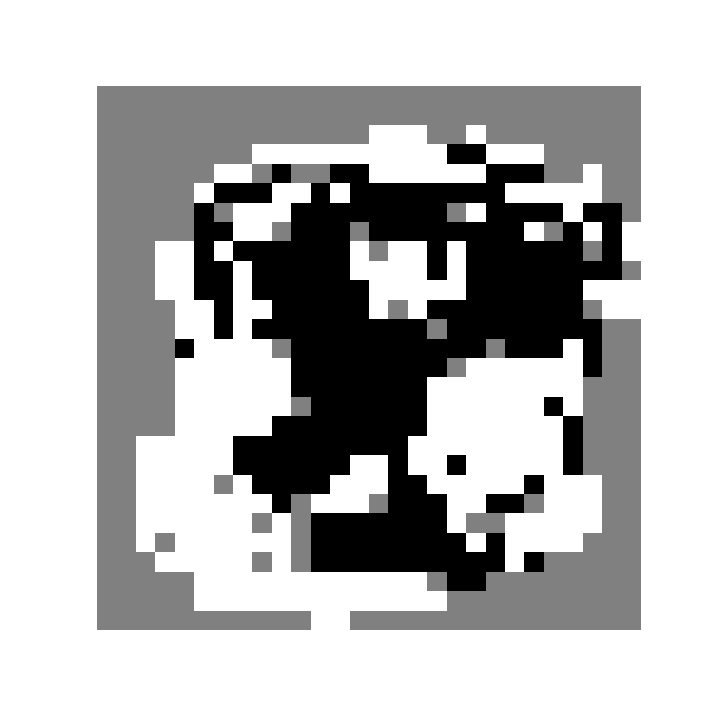}
        \caption{Number 8}
        \label{fig:image_8}
    \end{subfigure}
    \begin{subfigure}{.19\textwidth}
    \centering
        \includegraphics[width=\linewidth]{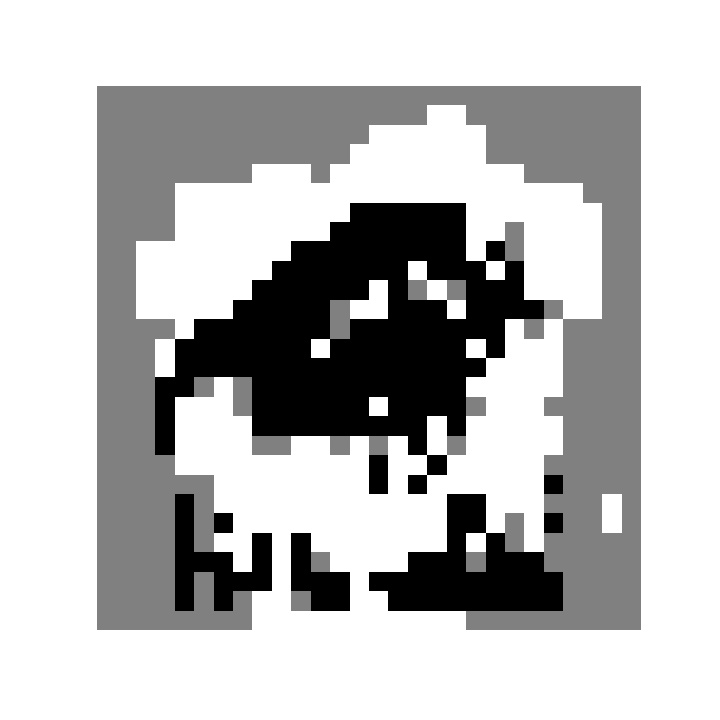}
        \caption{Number 9}
        \label{fig:image_9}
    \end{subfigure}
    \caption{Visualization of interpretability results for MNIST over each number 
$c\in C=\{0,\dots, 9\}$. A pixel is colored to: black if the value of decision variable 
$w^{+}_{f,c}$ is 1, white if the value of decision variable $w^{-}_{f,c}$ is 1, and 
gray otherwise. Intuitively, each subfigure $c$ visually represents what the learned 
binarized regression model predicts number $c$ to look like (and not to look like) 
based on its learned weights.}
  \label{fig:int_res}
\end{figure}

In order to demonstrate the simulatability of the learned binarized regression 
model, we simply visualize the learned non-zero weights of the learned binarized 
regression model as follows. For each class $c\in C=\{0,1,\dots, 9\}$, we color 
pixel $f\in F=\{0,1,\dots, 783\}$ to: black if the value of decision variable 
$w^{+}_{f,c}$ that is obtained from solving the MIP model is 1, white if the 
value of decision variable $w^{-}_{f,c}$ that is obtained from solving the MIP 
model is 1, and gray otherwise. Given the visualization of the learned weights 
in Figure~\ref{fig:int_res}, the users can take any input image and simulate 
the prediction of the learned model by visually comparing 
the input image to the subfigures~\ref{fig:image_0} to \ref{fig:image_9}. 
Intuitively, each subfigure $c$ visually represents what the learned 
binarized regression model predicts number $c$ to look like (and not 
to look like) based on its learned weights.

\section{Discussion, Related Work and Future Work} 
\label{sec:disc_rel}

In this section, we discuss our assumptions, choices, contributions 
and experimental results in relation to the literature with the goal 
of opening new areas for future work.

In Section~\ref{sec:mod_ac}, we detailed our model assumptions and 
choices for training accurate and interpretable binarized 
models using MIP and PBO. Similar to Rosenfeld et al.~\cite{Rosenfeld2020}, 
we made the assumption on the existence of function $\mathbb{g}$ which allowed 
us to train linear regression models for the multiclass classification task. 
In Section~\ref{sec:exp1}, we experimentally demonstrated that modeling 
$\mathbb{g}$ instead of $\mathbb{f}$ significantly improves the test 
performance of our MIP model. Our results suggest that similar works 
on training Binarized Neural Networks~\cite{Hubara2016} using MIP 
models~\cite{Icarte2019} might also benefit from similar explicit 
modeling of function $\mathbb{g}$. 

In Section~\ref{sec:exp2}, we experimentally demonstrated that 
our learned binarized regression model is robust to random label 
corruption. Under adversarial settings, the data corruption problem 
can be formulated as a bilevel optimization problem where the attacker 
tries to optimally corrupt the dataset in order to degrade the test 
accuracy of the machine learning model~\cite{Suvak2021}. Under such 
adversarial settings, the ability to provide \textit{formal} robustness 
guarantees~\cite{Natarajan2013} presents an important venue for 
future work.\footnote{We have experimented with a modified version 
of our MIP model that uses big-M constraints to remove outliers, and found 
that this model performed worse than our original MIP model. We conjecture 
that this is because (i) the balanced optimization of prediction 
margin and model size is sufficient for good experimental performance under 
our experimental settings and (ii) the modified MIP model is harder to 
optimize due to the big-M constraints.} Similar to this 
paper, the robustness study of other interpretable machine learning models 
(e.g., decision lists~\cite{Rivest1987}, decision sets~\cite{Clark1989}, 
decision trees~\cite{Breiman1984}) can potentially yield important ideas 
for future work.

\begin{figure}
\centering
    \begin{subfigure}{.24\textwidth}
    \centering
        \includegraphics[width=\linewidth]{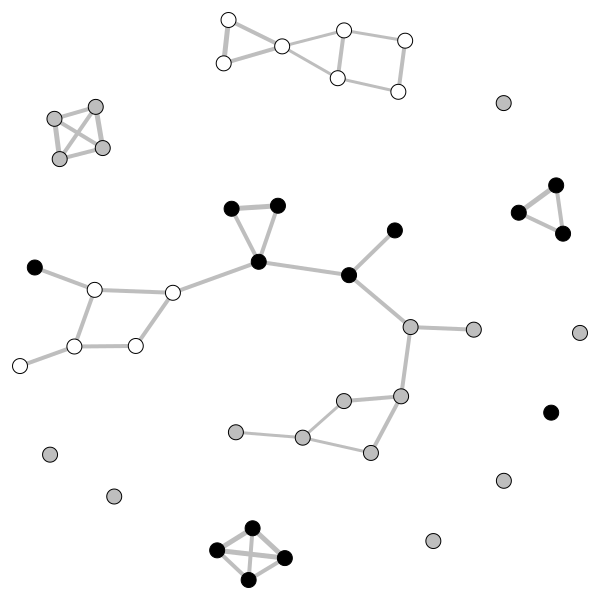}
        \caption{Flags 0}
        \label{fig:flags_class_0}
    \end{subfigure}
    \begin{subfigure}{.24\textwidth}
    \centering
        \includegraphics[width=\linewidth]{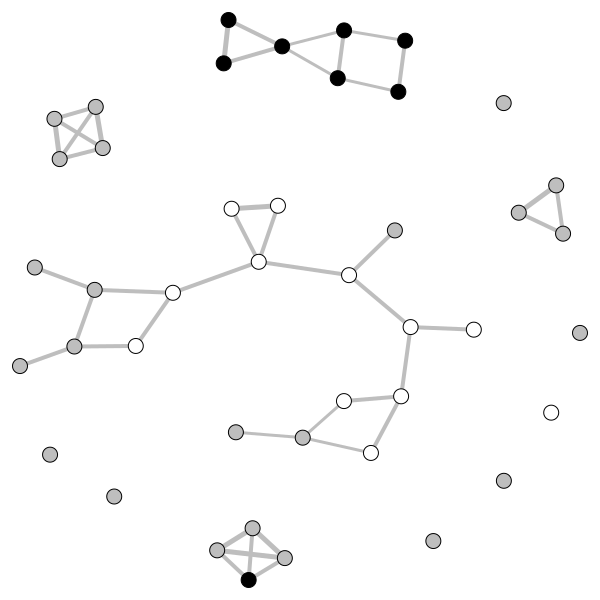}
        \caption{Flags 1}
        \label{fig:flags_class_1}
    \end{subfigure}
    
    \begin{subfigure}{.24\textwidth}
    \centering
        \includegraphics[width=\linewidth]{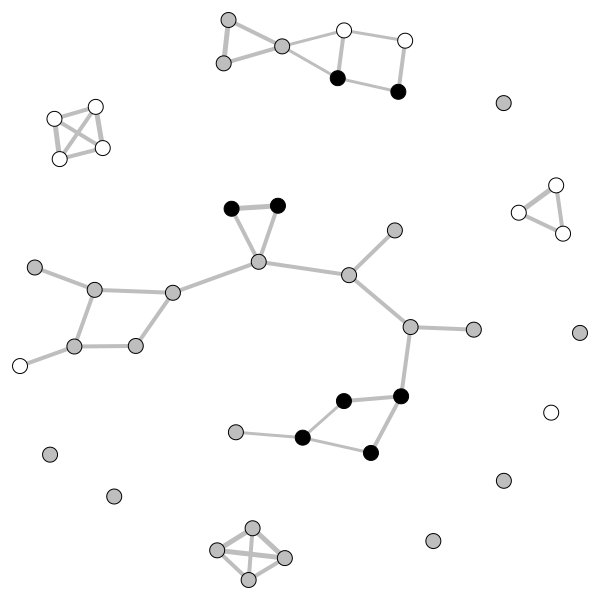}
        \caption{Flags 2}
        \label{fig:flags_class_2}
    \end{subfigure}
    \begin{subfigure}{.24\textwidth}
    \centering
        \includegraphics[width=\linewidth]{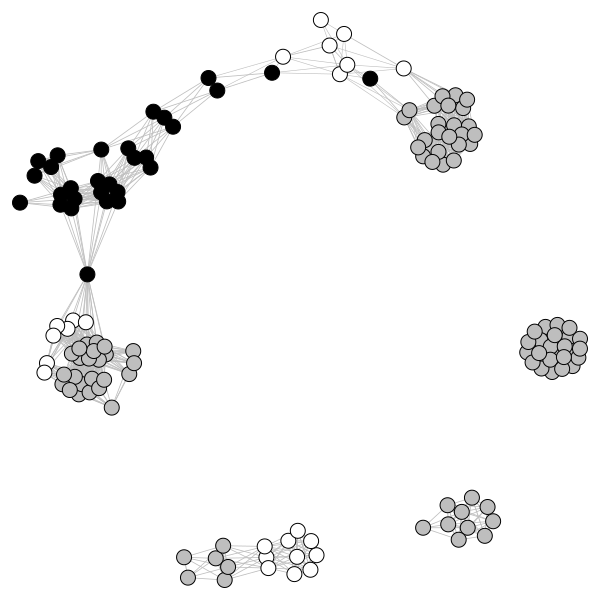}
        \caption{Ubuntu 0}
        \label{fig:ubuntu_class_0}
    \end{subfigure}
    
    \begin{subfigure}{.24\textwidth}
    \centering
        \includegraphics[width=\linewidth]{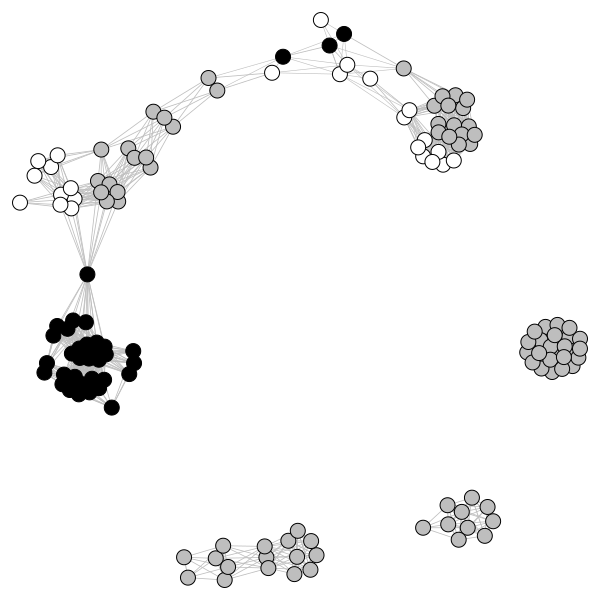}
        \caption{Ubuntu 1}
        \label{fig:ubuntu_class_1}
    \end{subfigure}
    \begin{subfigure}{.24\textwidth}
    \centering
        \includegraphics[width=\linewidth]{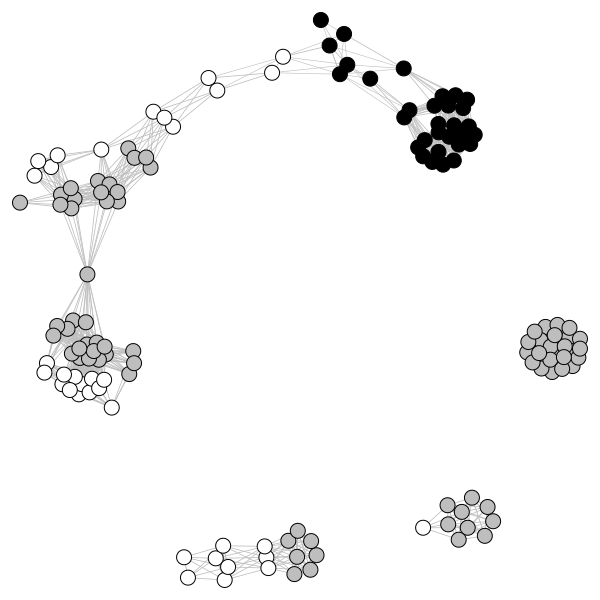}
        \caption{Ubuntu 2}
        \label{fig:ubuntu_class_2}
    \end{subfigure}
    \caption{Visualization of interpretability results for Flags and Ask Ubuntu 
datasets over different classes using the same coloration methodology that is 
described in Figure~\ref{fig:int_res} where each vertex represents a feature 
and two vertices are connected by an edge based on their Hamming distance to 
each other. Here, we have used the Continuous k-Nearest Neighbours 
method~\cite{Berry2019} such that an edge is defined for a pair of features 
$x,y\in F$ if $d(x,y) < \delta * \sqrt{d(x,x_k) d(y,y_k)}$ where $d(x,y)$ 
denotes the Hamming distance between features $x$ and $y$, $\delta \in \mathbb{R_+}$ 
is a parameter controlling the sparsity of the graph, and features 
$x_k, y_k \in F$ are the k-th nearest neighbours of features $x$ and $y$, respectively.}
  \label{fig:int_res_2}
\end{figure}

In Section~\ref{sec:int_res}, we experimentally demonstrated the 
interpretability of our learned binarized regression model on 
a challenging visual task (i.e., MNIST). The effective 
visualization of our learned models on \textit{non-visual} tasks 
(e.g., as visualized in subfigures \ref{fig:flags_class_0}-\ref{fig:flags_class_2} 
and \ref{fig:ubuntu_class_0}-\ref{fig:ubuntu_class_1} for Flags and Ask Ubuntu 
datasets) in order to derive valuable insights remains an important 
area for future work.

\bibliographystyle{IEEEtran}
\bibliography{IEEEfull,mybibliography}

\end{document}